\documentclass[runningheads]{llncs}
\usepackage{graphicx}
\usepackage{tikz}
\usepackage{comment}
\usepackage{amsmath,amssymb} % define this before the line numbering.
\usepackage{color}
\usepackage{subcaption}
\usepackage{multirow}

\begin{document}
% \linenumbers
\pagestyle{headings}
\mainmatter
\def\ECCV16SubNumber{***}  % Insert your submission number here

\title{Recovering Detail in 3D Shapes Using Disparity Maps} % Replace with your title

\titlerunning{Recovering Detail in 3D Shapes}
\author{Marissa Ramirez de Chanlatte \inst{1}\and
Matheus Gadelha\inst{2} \and
Thibault Groueix\inst{2} \and
Radomir Mech\inst{2}}
\authorrunning{M. Ramirez de Chanlatte et al.}
\institute{University of California, Berkeley 
 \and
Adobe Research}

\maketitle

\begin{abstract}
We present a fine-tuning method to improve the appearance of 3D geometries reconstructed from single images. We leverage advances in monocular depth estimation to obtain disparity maps and present a novel approach to transforming 2D normalized disparity maps into 3D point clouds by using shape priors to solve an optimization on the relevant camera parameters. After creating a 3D point cloud from disparity, we introduce a method to combine the new point cloud with existing information to form a more faithful and detailed final geometry. We demonstrate the efficacy of our approach with multiple experiments on both synthetic and real images. 
\end{abstract}

\section{Introduction}
% \begin{figure*}
%     \centering
%     \includegraphics[width=.8\textwidth]{Figures/visual_abstract.png}
%     \caption{Our method projects the output of a monocular depth estimator to 3D, using it to improve the appearance of a 3D geometry from a single view reconstruction model.}
%     \label{fig:method}
% \end{figure*}

Reconstructing full 3D shapes from single RGB images is a longstanding computer vision problem.
A system capable of performing such task needs to have some understanding of 
\emph{pixels} -- how the scene illumination interacts with the geometry and materials to create that
image -- and ~\emph{shapes} -- their structure, how they are usually decomposed and 
which symmetries arise from their function and style.
To develop such a system, a learning-based procedure would require an abundance of
\emph{paired} data in those two forms.
Unfortunately, a practical limitation comes from the fact that we have significantly
more access to images (pixels) than 3D models (shapes).

\begin{figure}[h]
    \centering
    \includegraphics[width=\textwidth]{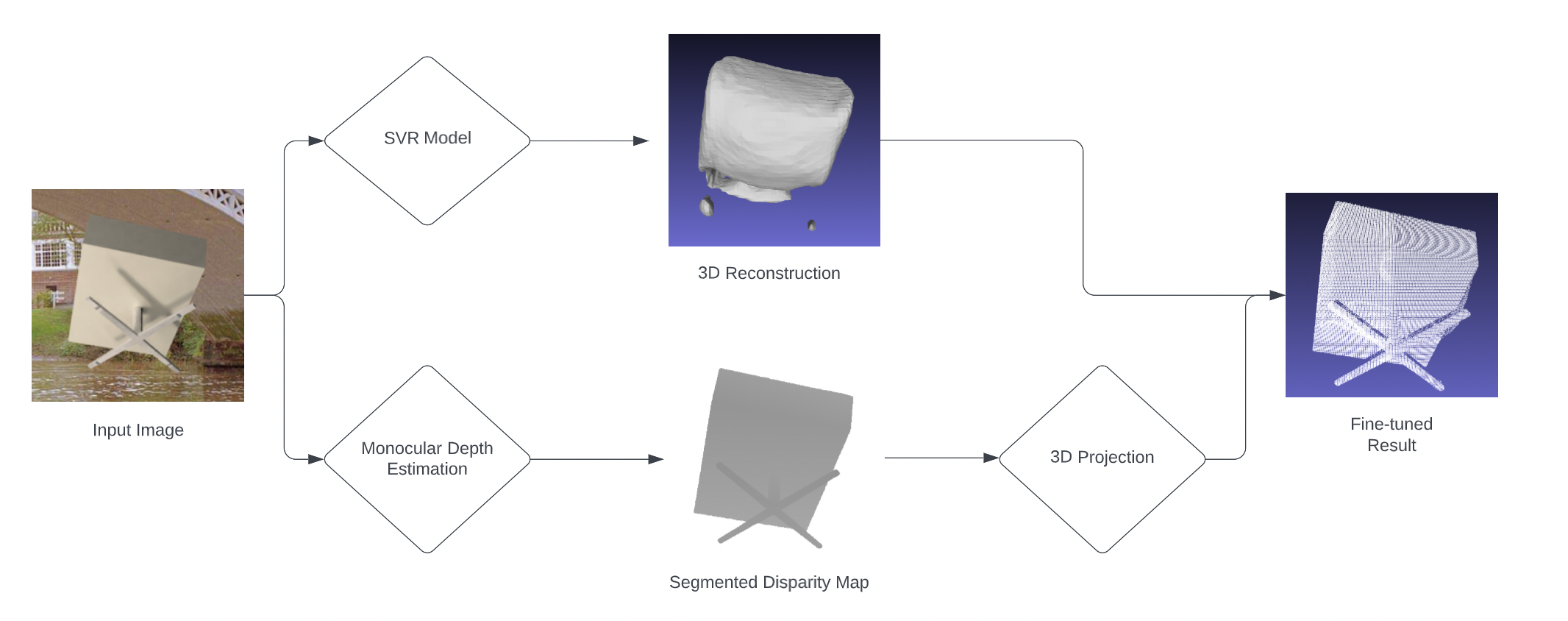}
    \caption{Our method projects the output of a monocular depth estimator to 3D, using it to improve the appearance of a 3D geometry from a single view reconstruction model.}
    \label{fig:method}
\end{figure}

There are many learning-based techniques that encode knowledge about \emph{shapes} \cite{choy20163d,autosdf2022,3dgan}. Most of them are trained using data from ShapeNet~\cite{shapenet2015}, a 3D mesh dataset with
tens of thousands of shapes.
Models trained using this data can perform various conditional shape generation tasks, like shape interpolation, completion and reconstruction from single views. Most of the approaches trained on ShapeNet have only a very rudimentary knowledge of pixels from real images.
While we are able to generate images using 3D meshes (i.e. rendering), those images are very different
from the ones we capture from our cameras in the real world.
Despite the rich behaviors that can be simulated through computer graphics, the research community
is far from recreating the abundant visual experience we encounter in our lives.
On the other hand, monocular depth estimation models are capable of inferring the partial
scene geometry quite well, even on real images.
Unlike models trained on ShapeNet, depth estimation techniques employ a variety of training data -- synthetic scenes, scanners, stereoscopic images, structure from motion, and so on.
The variety of training data leads to models capable of performing the task
remarkably well on a variety of scenarios including real images. %They can extract partial geometry from \emph{real} images -- they understand~\emph{real pixels} much better than approaches trained solely on synthetic ShapeNet objects.

The central proposition of this work is to use the knowledge of \emph{pixels} from depth estimation
approaches to improve models estimating \emph{shapes} from single RGB images.
A key challenge lies in the fact that, despite their name, monocular depth estimation techniques
actually predict disparity, the distance between two points in a stereo image pair, rather than depth. \emph{In order to recover true depth, one still needs to estimate
intrinsic camera parameters and solve scale and distance ambiguities}. This small but important step has been understudied.
Whereas other works have attempted to convert disparity information to depth~\cite{Yin_2021_CVPR}, they are restricted to visible portions
of full scenes and do not leverage any knowledge of particular shapes.
Our solution, which to our knowledge is the first to address converting disparity to depth on single shapes, is to use shape estimates coming from vanilla single-view reconstruction models to recover image parameters that align the disparity estimations with the object coordinate system.
The geometry extracted from the disparity estimation models can then be used to refine 3D shapes
from single-view reconstruction techniques.
Notably, our method can be applied to any view-centric single-view reconstruction technique without
needing extra training or additional data, unlike data-driven approaches such as~\cite{Yin_2021_CVPR}.
Similarly, as monocular depth estimation models get better, their improvements can be immediately
translated into their single-view reconstruction counterparts.
Finally, we demonstrate the efficacy of our technique in experiments with both synthetic and real images.

\section{Related Work}

\paragraph{Single-view 3D reconstruction (SVR).}

Computational approaches to SVR date back to at least Roberts' PhD thesis~\cite{roberts1963machine} in 1963, and the problem was already known to be ill-posed by the scientist Alhazen in the 11th century. Most related to our work are recent deep learning approaches which currently present state-of-the-art results for this problem.

\noindent \textit{Deep Networks for 3D Shape Representation.}
The choice of 3D data representation is one of the main discriminating factors to analyze deep learning approaches for SVR.  % and is still arguably an open question, even though there is a trend in recent work towards generating implicit functions.
% Choy et al.~\cite{choy20163d} propose to use 3D voxel-grids, as a natural extension to 2D pixel grids, on which to perform 3D convolution~\cite{autosdf2022,3dgan}. 
% % , and several paper have explored using 3D convolutions on such grid to perform analysis and generation. 
% The drawback of this representation is that the memory consumption scales cubically with respect to the resolution, making it hard to model fine details, though subsequent work introduced optimized grid structures~\cite{Hane:2017,Riegler2017THREEDV,TDB17b}, and other grid-based approaches from the NeRF community~\cite{mildenhall2020nerf} use tensor decomposition~\cite{TensoRF} and hash tables~\cite{mueller2022instant,yu_and_fridovichkeil2021plenoxels}.
Choy et al.~\cite{choy20163d} propose a volumetric representation using 3D voxel-grids~\cite{autosdf2022,3dgan}, as a natural extension to 2D pixel grids on which to perform 3D convolutions. 
Several solutions have been explored to mitigate the prohibitive memory consumption of using 3D grids in deep learning.
Previously, these solutions were focused on occupancy grids  ~\cite{Hane:2017,Riegler2017THREEDV,TDB17b}, but several more recent radiance fields alternatives using hash tables and tensor decomposition ~\cite{mildenhall2020nerf,TensoRF,mueller2022instant,yu_and_fridovichkeil2021plenoxels} were also proposed. 
Instead of using grids, three papers \cite{park2019deepsdf,mescheder2019occupancy,chen2019learning} concurrently pioneered the use coordinate-based Multi-Layered Perceptrons to model a 3D volume~\cite{thai20213d}. 
Several other approaches model a surface instead of a volume. Fan et al.~\cite{Fan:2017:cvpr} pioneered an approach to generate point clouds on a surface~\cite{mrt18,Luo_2021_CVPR}. 
Other techniques model 3D surfaces via parametric deformations from a reference surface~\cite{groueix2018,wang2018pixel2mesh,gkioxari2019mesh}. 
Despite a lot of progress in representing shapes, all the aforementioned SVR methods suffer when applied to real images.
    Our approach can be thought of as a modular extension to some of these methods that
    leverages the knowledge of monocular depth estimation techniques, which can be trained on real images,
    thus generalizing better to more realistic scenarios.
% , and several paper have explored using 3D convolutions on such grid to perform analysis and generation. 

% , and demonstrate state-of-the-art results in some generative tasks~\cite{autosdf2022,3dgan}.

% . Several approaches have optimized this representation with octrees [chaene, tatar...], and they have proved to achieve compelling results in generative tasks [3D-GAN, AutoSDF]

% \begin{itemize}
%     \item Tusliani and Malik seminal SVR.
%     \item Voxel-based approaches as natural extension to 2D convs (Vox3D, 3D-GAN, AutoSDF).
%     \item Point clouds (MRTNet, view-based, Su et. al.)
%     \item Parametric deformations (AtlasNet, Pixel2Mesh, MeshRCNN -- this is hybrid).
%     \item Neural fields (OccNet, DeepSDF, 3DShapeGen, IMNet).
%     \item Takeaway: a lot of progress on representing shapes, nowadays we are capable of encoding diverse
%     shapes with good level of detail; however, all those methods suffer when applied to real images.
%     Our approach can be thought of a modular extension to some of these methods that
%     leverages the knowledge of monocular depth estimation models, which can be trained on real images,
%     thus generalizing better to more realistic scenarios.
% \end{itemize}

\noindent\textit{Single-view 3D reconstruction from real images.}
There are two main strategies in SVR to target performance on real images. The first one is through the training data \textit{i.e.} to have real images in the training set. Existing datasets  with associated ground truth 3D model have clear limitations.
Pix3D~\cite{sun2018pix3d} does not have good diversity in terms of shapes (only about 700) and mostly contains furniture.
ObjectNet~\cite{Barbu2019ObjectNetAL} and Pascal3D~\cite{xiang_wacv14} don't have good image-shape alignment. Some approaches aim to learn SVR using image-only datasets using differentiable reprojection losses~\cite{vasudev2022ss3d,umr2020,cmrKanazawa18,ucmrGoel20,tulsiani2020implicit,lin2020sdfsrn,ye2021shelf}, though this is an extremely challenging task. Another strategy is to use domain adaption techniques to bridge the gap between synthetic renderings and real images, for instance by imposing depth and normal as an intermediate representation between RGB images and the full 3D geometry~\cite{marrnet,shapehd,genre,shin2018pixels}. These techniques show some improvement on real images but within the scope of the categories spanned by the 3D dataset. They thus don't fully leverage the generality of monocular depth estimation methods.
Whereas those approaches perform better than vanilla models trained only on ShapeNet or Pix3D,
    they still fail to generalize to real scenarios.
    Our method is complimentary to these and can be used along monocular depth estimation in addition to
    domain adaptation techniques.
% \begin{itemize}
%     \item Pascal3D, ObjectNet, Pix3D. Pix3D does not have good diversity in terms of shapes (only about 700); ObjectNet and Pascal3D don't have good image-shape alignment.
%     \item Some approaches rely on domain adaptation (Domain Adaptation SVR Paper) others in intermediate
%     representations (GenRe, MarrNet, follow up to MarrNet).
%     \item Whereas those approaches perform better than vanilla models trained only on ShapeNet or Pix3D,
%     they still fail to generalize to real scenarios.
%     Our method is complimentary to these and can be used along monocular depth estimation in addition to
%     domain adaptation techniques.
% \end{itemize}

\paragraph{Monocular depth estimation.}

In their seminal work make3D, Saxena et al.~\cite{saxena} cast depth estimation as a supervised learning problem trained on a dataset of laser scans. 
Several subsequent papers have improved the architectures~\cite{Eigen2,laina2016deeper,roy2016monocular,Depth2015CVPR,li2018deep}, the losses~\cite{Eigen,depthclassif,ordinal} and post-processing steps~\cite{crfdepth,Miangoleh2021Boosting}. 
We organize our discussion around the training data since the robustness and generality of monocular depth estimation approaches stems from the diversity of the training datasets~\cite{Ranftl2020}. 
% sensors
Laser scanners~\cite{schoeps2017cvpr,6248074,book} based on time-of-flight as well as sensors based on structured light~\cite{Khoshelham2012AccuracyAR,7251485} provide ground truth depth but sparse annotations for dynamic scenes.
% SfM
Structure-from-motion can also be used to obtain sparse 3D ground-truth, up to scale, from multi-view images of a static scene~\cite{MDLi18}. 
% stereo 
Garg et al.~\cite{garg2016unsupervised} propose to use rectified stereo pairs as supervision~\cite{monodepth17,monodepth2,Luo2018SVS} but the corresponding datasets are not all calibrated, providing disparity up to scale and shift.
% ordinal
Chen et al. created a dataset where ordinal relationships between pixels are manually annotated~\cite{NIPS2016_0deb1c54}.
% Toward robust
MidasNet~\cite{Ranftl2020} pioneered leveraging those diverse sources of data by estimating normalized disparity and achieving  breakthrough results in generalization capabilities of monocular depth estimation. 
% takeaway
Recent monocular depth estimation models~\cite{Yin_2021_CVPR,Miangoleh2021Boosting} show remarkable performance in various scenarios. 
Those models can capture the visible part of the scene geometry but have little knowledge about shapes. 
For example if three legs of a chair are visible, the depth estimation models will not be very helpful for completing the shape
and generating the remaining leg.
On the other hand, SVR models might miss fine details of visible parts of the chair that were captured by the depth estimation network, but
will most certainly create a four-legged chair because it is capable of reasoning about the overall shape structure -- it was supervised using complete chair meshes during training.
Our goal is to try to get the best of both worlds: more details when reconstructing shapes from real images while maintaining the
coarse estimation from SVR models.

% \begin{itemize}
%     \item Some old work on depth from stereo (see towards robust depth estimation paper).
%     \item Deep neural networks for monocular depth estimation and datasets
%     \item MidasNet (Towards monocular depth estimation), using datasets from different sources (stereo, scanners)
%     \item New architectures (Jianming, dense transformers)
%     \item Takeaway: monocular depth estimation models are getting more accurate show remarkable
%     performance in various scenarios. Those models can capture the visible part of the scene geometry but
%     have little knowledge about shapes.
%     Our goal is to use those models to improve shape reconstruction from real images, endowing models
%     trained to represent tens of thousands of shapes with an improved ability to recover shapes
%     from real images.
% \end{itemize}

\section{Method}
We start with an off-the-shelf and pre-trained single-image predictor that produces an unsatisfactory result and aim to fine-tune that result to better match the input image. Take for example an image of a chair. The network may produce a geometry that looks like a chair, but perhaps the shape of the legs or detail in the back is inaccurate. Our method takes that initial prediction and uses a disparity map to improve the appearance of the geometry. 

The fine-tuning procedure is as follows:

\begin{enumerate}
    \item Get a point cloud, $\Lambda$, representation of the initial prediction. 
    
    \item Split $\Lambda$ into two sub-pointclouds of points that are visible, $\Lambda_{vis}$, and occluded, $\Lambda_{occ}$, from the camera viewpoint of the input image. 
    \item Given a 2D disparity map of image $I$ called $I_{disp}$ find the focal length, scale, displacement, and translation constants, $\omega = \{fov, s, t, z_t\}$, necessary to project $I_{disp}$ into $\Psi$, the 3D point cloud most closely resembling $\Lambda_{vis}$.
    \item Combine $\Lambda_{occ}$ and $\Psi$ and remove any points from $\Lambda_{occ}$ that are now visible. Call this cleaned and combined final point cloud $\Psi'$.
\end{enumerate}

In step 1, we start with a point cloud representation of the initial prediction of our object. This point cloud can come from any geometry including a mesh or implicit function. We are not limited to networks that produce any particular type of geometry. 

After we have our initial prediction, we split the point cloud into visible and occluded points. The idea here is to leverage the information we have from the disparity map while keeping the points for which we have no disparity  information (the occluded points) the same. 

In step 3, we take the predicted 2D disparity map and convert it into a 3D point cloud that will replace the visible part of the object. This is the most technically challenging step and our key contribution. Because of its importance we will discuss the details of this step further in the next section

The final step is to combine the occluded point cloud, $\Lambda_{occ}$ from the initial prediction with the projected disparity points, $\Psi$. Then a final cleaning is performed, to remove any occluded points that are no longer occluded by the new visible geometry. This is done by iteratively removing points from $\Lambda_{occ}$ that are not occluded by the projected disparity points, $\Psi$. This is to account for errors in the initial prediction. Imagine looking at the back of a chair that is mostly open, connected only by spokes. If the initial prediction erroneously reconstructed the chair with a solid back, replacing the visible points with the projected disparity map should restore this detail, but only on the visible side of the back of the chair. The other side will still be solid. We must also remove extraneous points from the original set of occluded points, $\Lambda_{occ}$. 
The final result is the point cloud $\Psi'$. For simplicity, we leave the geometry as a point cloud in the experiments, but it can easily be transformed into a mesh~\cite{psr}  or an implicit function~\cite{NeuralPull} as desired because $\Psi'$ is dense. This meshing step will also get rid of any discontinuity between the previous occluded points and the new visible points added from the depth map. 

\subsection{Disparity Projection}
The heart of this method is the disparity projection. In the literature, people often discuss the task of ``depth" prediction of a single image, but in most cases what is being predicted is disparity, or the difference between two matching points in two stereo images. Disparity is related to inverse depth by an affine transformation and is used in monocular depth estimation \cite{wang2019web,ranftl2020towards, Yin_2021_CVPR} for several reasons. The first being the availability of disparity data and the difficulty of converting disparity to depth. In mixed datasets it is easier to convert depth maps to disparity maps than the other way around, so disparity maps are popular training sets. Second, because of its relationship to inverse depth, using a disparity map for training effectively weights the loss function by depth, biasing towards objects in the foreground. 

Without knowledge of several camera parameters, it is impossible to get true depth from normalized disparity. We assume we are starting with a normalized disparity map $d$ which is related to true depth, $Z$, through the following relationship:
\begin{equation}
    \frac{1}{Z} = \frac{d - (c_x^R - c_z^L)}{fb}, 
\end{equation}
where $b$ is the camera baseline, $f$ is the focal length, and $c_x^R, c_x^L$ are the horizontal coordinates of the principal points from the stereo camera pair. For simplicity, we rewrite this equation as
\begin{equation}
    \frac{1}{Z} = s*d + t,
\end{equation}
where $s=\frac{1}{fb}$ and $t=-\frac{(c_x^R - c_z^L)}{fb}$. 

These equations are used to convert disparity to depth, but then depth coordinates $(u, v)$ must be converted to 3D coordinates $(X, Y, Z)$ to produce the point cloud necessary for our method. To do this, we invert the perspective projection equation in the pinhole camera model:
 
\begin{equation}
    \begin{bmatrix}
    u \\
    v \\
    1
    \end{bmatrix} = \frac{1}{Z}
    \begin{bmatrix}
    f_x & 0 & u_0 \\
    0 & f_y & v_0 \\
    0 & 0 & 1
    \end{bmatrix}
    \begin{bmatrix}
    X \\ 
    Y \\ 
    Z \\
    \end{bmatrix} 
\end{equation}
where $(f_x, f_y)$ is the focal length and $(u_0, v_0)$ is the optical center of the image plane. 

Solving for $(X, Y, Z)$ we get

\begin{align}
    u = \frac{f}{Z}X + u_0 \implies X = \frac{(u - u_0)}{f_x}Z\\ 
    v = \frac{f}{Z}Y + v_0 \implies Y = \frac{(v - v_0)}{f_y}Z
\end{align}
 where $f$ is the focal length defined as $f=\frac{x}{2tan(fov/2)}$ and $fov$ is the field of view and $x$ is the diagonal of the image. We assume $(u_0, v_0)$ to be $(0, 0)$. When we are given only a normalized disparity map, $s, t,$ and $fov$ are all unknown, so we must find acceptable approximations of these parameters in order to get the desired 3D point cloud. 

The pinhole camera model assumes the origin of the coordinate system to be the camera position, while most view-centered reconstruction models assume the reconstructed geometry to be placed at the origin. To relate the two coordinate systems, we need to predict an additional parameter, a translation constant $z_t$. We translate our projection to get $Z' = Z + z_t$ where $Z'$ is aligned with the initial object reconstruction. In total this gives a set four parameters, $\omega = \{s, t, fov, z_t\}$ that we must predict. 
%Note that if the single-image predictor were object-centric, we would have to predict a rotation to relate a view-centric coordinate-system with an object centric one. Predicting this rotation is subject to local minima. In cases where a canonical coordinate system is important, we argue that it is better to learn this rotation during training of the single-view predictor rather than optimize it in post-processing, since learning on a data collection might help alleviate these minima from the loss landscape.
    
Combined, the equations above form a function $f(s, t, fov, z_t, d)=\Psi$ which takes a normalized disparity map along with other parameters and transforms it into the corresponding 3D point cloud, $\Psi$. To approximate the parameters $\omega$, we fix the disparity map, $d$, and use stochastic gradient descent to minimize the Chamfer distance between $\Psi$ and the initially predicted visible points $\Lambda_{vis}$.

\begin{align}
    loss = \frac{1}{|\Psi|}\sum\limits_{x\in \Psi} &\min_{y \in \Lambda_{vis}} || x - y||_2^2 + \label{eq:loss} \\ &\frac{1}{|\Lambda_{vis}|}\sum\limits_{y\in \Lambda_{vis}} \min_{x \in \Psi} || x - y||_2^2 \nonumber
\end{align}

The parameters that minimize Eq.~\ref{eq:loss} are used to create the final point cloud used throughout the rest of the fine-tuning procedure. 

The intuition for the approach comes from the assumption that the initial prediction is roughly the right general shape and size as the desired object, and is only lacking in detail. By fitting the disparity map to the initial visible points, we should recover camera parameters that lead to a point cloud of roughly the right shape.

It is important to note that this is an overdetermined function. There are infinite choices of $\omega$ that will result in the same point cloud. There are also many local minima. This makes this approach very sensitive to initialization. To improve robustness, we perform the fitting several times, each time randomly initializing, and then choosing $\omega$ that resulted from initialization with best loss as the final parameters. 
\section{Experiments}
To validate our method, we perform experiments on two different off-the-shelf reconstruction networks that were trained on two different data sets. The first dataset is a synthetic dataset formed by rendering ShapeNet objects in different poses and lighting conditions. The second is Pix3d, a real-world data set. In both cases we focus on the category of chairs. 

We evaluated the final result using two metrics: f-score and Chamfer distance. 

\begin{figure*}
    \centering
    \includegraphics[width=.45\textwidth]{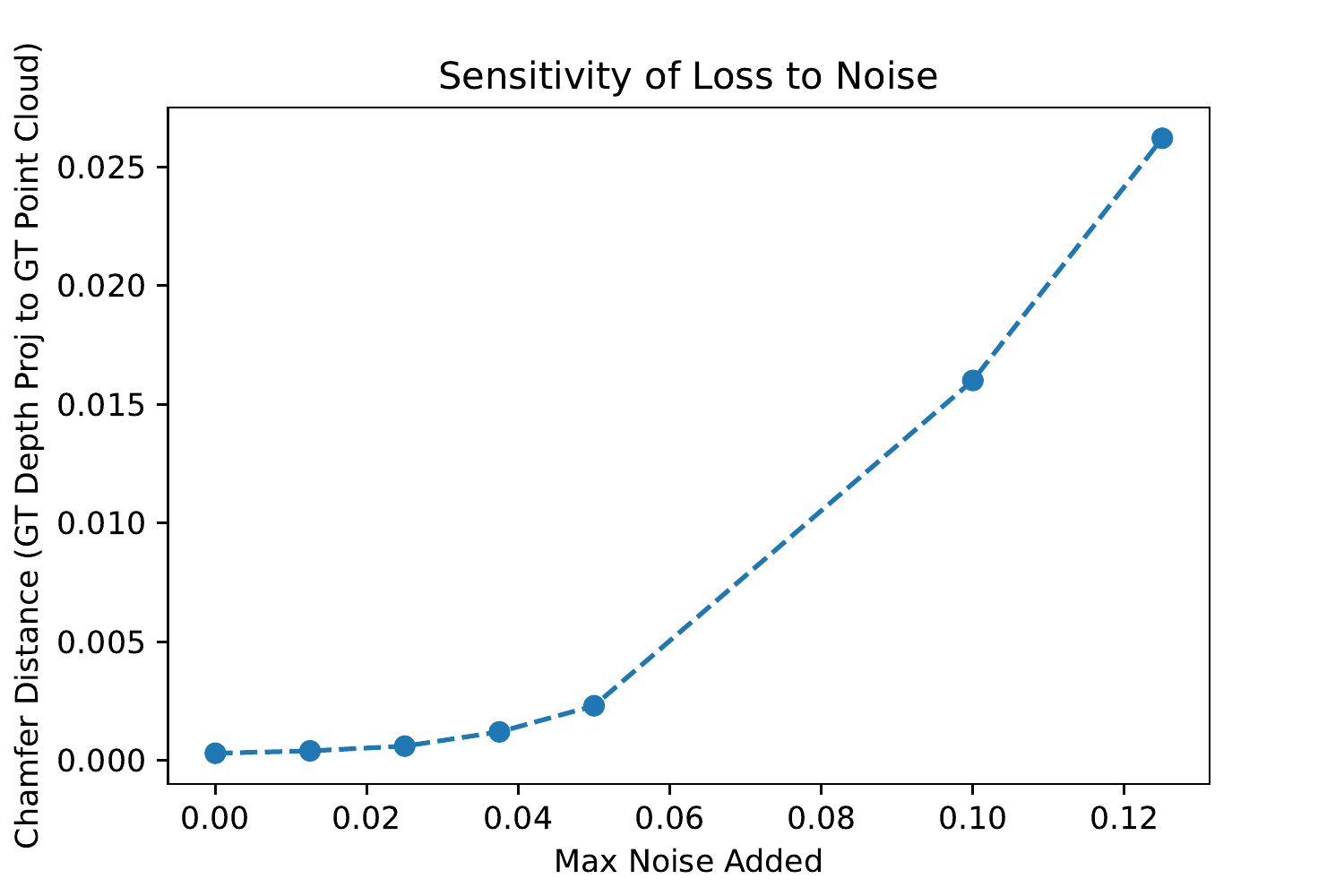}
    \includegraphics[width=.45\textwidth]{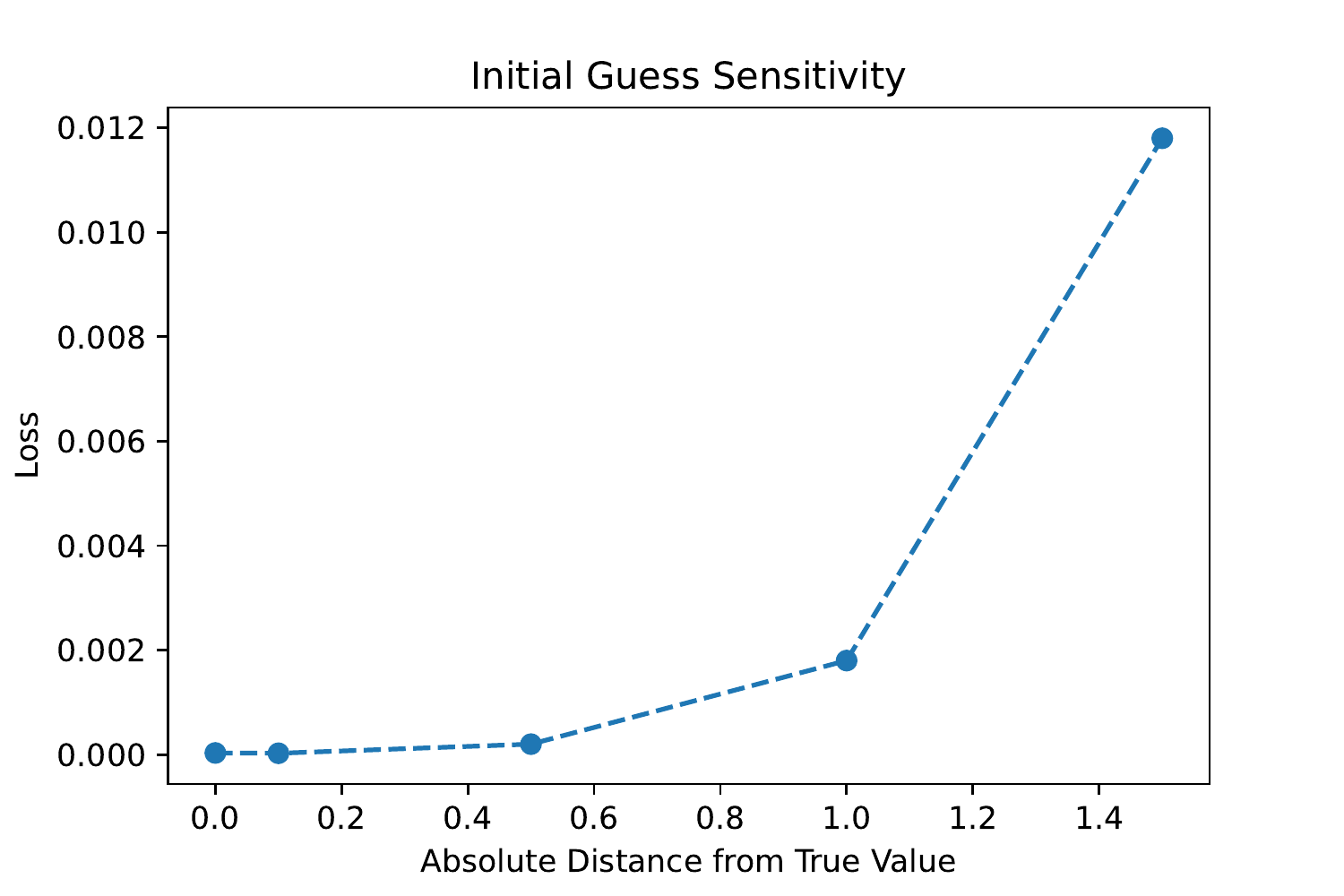}
    \caption{\textbf{Left:} our 3D projection is robust to some amount of noise in the input disparity map, but deteriorates if the noise level goes beyond $0.05$.  \textbf{Right:} the error in the projection grows as a function of the distance between the initial guess and the true camera parameters, because of local minima in our optimization problem. We adress this issue by randomly choosing multiple initializations.}
    \label{fig:noise_sensitivity}
\end{figure*}

\subsection{Sensitivity Studies}
Before discussing full-scale experiments, we first present two studies to examine the sensitivity of our method to errors in the disparity map and the initialization of parameters. 

To examine the effect of error in disparity prediction, we simulate error by uniformly adding noise to ground truth disparity maps. Using our disparity projection method, we estimate camera parameters. We then use those camera parameters to project the ground truth disparity and measure the Chamfer distance from the ground truth object. As can be seen in Fig.~\ref{fig:noise_sensitivity}, the Chamfer distance stays pretty small for small amounts of noise, but as more noise is added, the Chamfer distance increases, indicating some level of sensitivity to errors in prediction. However, it is important to remember that errors in disparity prediction are often not uniform. The more common failure mode is that errors are concentrated in one area (i.e. the back leg predicted to be closer than it actually is, or fading into the background completely). 

The second case of sensitivity we examine is sensitivity to initialization. To measure this we pick a case in which the initial parameters are known and perturb the initial guess steadily further away from the known value.  As can be seen in Fig.~\ref{fig:noise_sensitivity},  the final loss grows exponentially with distance from the initial value, indicating a significant sensitivity. This finding motivates the random initialization seen in our algorithm. We run the fitting multiple times and pick the final result with the lowest lost.

% \begin{figure}
%     \centering
%     \includegraphics[width=.45\textwidth]{Figures/noise_sensitivity.pdf}
%     \caption{As noise is added to the disparity map, error in the projection resulting from the predicted parameters increases as well, indicating a sensitivity to errors in the disparity map.}
%     \label{fig:noise_sensitivity}
% \end{figure}

% \begin{figure}
%     \centering
%     \includegraphics[width=.45\textwidth]{Figures/initialization_sensitivity.pdf}
%     \caption{The final parameter fitting loss as a function of the distance of the initial guess from the true value. Loss grows with distance indicating a sensitivity to the initial guess which we combat by randomly choosing multiple initializations.}
%     \label{fig:initial_guess}
% \end{figure}

\subsection{Synthetic Images}
\begin{figure}[h]
    \centering
    \begin{subfigure}{.45\textwidth}
    \resizebox{\textwidth}{!}{
     \begin{tabular}{|c|c|c|c|}
        
        \hline
        Method & & f-score & CD \\ 
        \hline
        \multirow{3}{*}{Baseline (3DShapeGen)}& Min. & 0.0577 & 0.0239 \\
        & Max. & 0.5653 & 0.2629 \\
        & Mean & 0.2250 & 0.1086 \\
        \hline
        \multirow{3}{*}{GT Depth}& Min. & 0.2639 & 0.0156 \\
        & Max. & 0.7824 & 0.1702 \\
        & Mean & 0.4938 & 0.0640 \\
        \hline
        \multirow{3}{*}{GT Disparity}& Min. & 0.0331 & 0.0216 \\
        & Max. & 0.7060 & 0.2444 \\
        & Mean & 0.2827 & 0.0944 \\
        \hline
    \end{tabular} }
    \end{subfigure}
    \begin{subfigure}{.5\textwidth}
        \includegraphics[width=\textwidth]{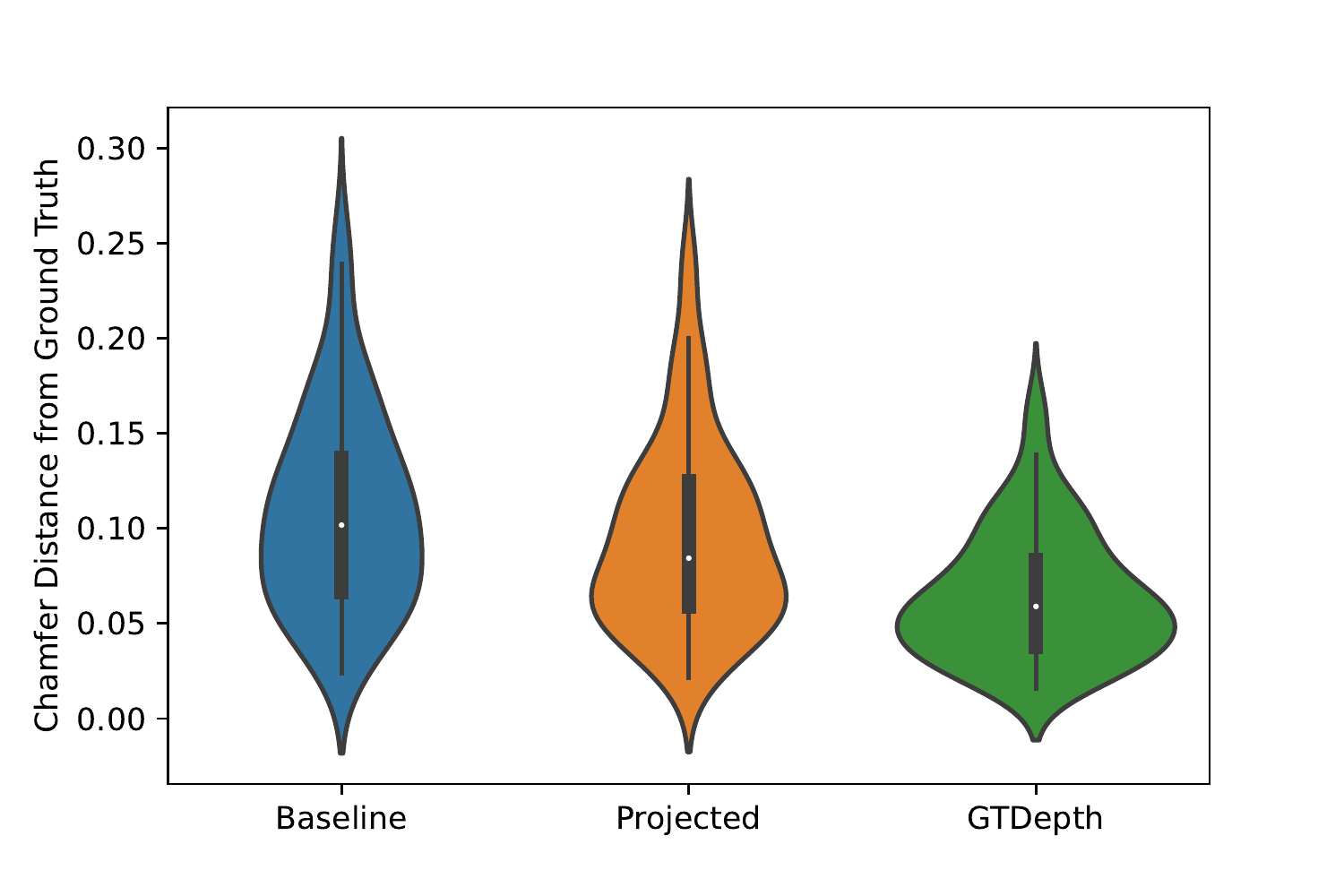}
    \end{subfigure}

    \caption{We compare a baseline SVR model, 3DShapeGen~\cite{thai20213d}, with two oracle versions of our approach using ground truth depth, and projected ground truth disparities, on synthetic renderings of ShapeNet chairs. Using ground truth depth yield a 180\% improvement in the f-score@1 in average. Despite not knowing camera parameters perfectly, our method still provides a 32\% improvement in the f-score@1 in average.}
    \label{tab:synth_results}
\end{figure}
We present two oracle experiments on synthetic data to clarify the sources of errors in our approach. 
In this section,  we use a dataset of renderings of ShapeNet~\cite{shapenet2015} objects from 3DShapeGen authors~\cite{thai20213d}. We use their image-only network to perform single-image reconstruction. 

First, we try merging the visible points of the ground truth to the 101 initial predictions provided by 3DShapeGen~\cite{thai20213d}. This achieves the maximum benefit possible with a perfect disparity map and a perfect projection of this map to 3D. On average, each object saw an average improvement of 180\% in the f-score at 1 and a 40\% improvement in Chamfer distance. This result demonstrates the potential for depth maps to improve the overall appearance and detail of a reconstructed object by an SVR method. 

As was discussed in section 3.1, the challenge is that in most cases, monocular depth estimation does not predict absolute depth maps, but instead normalized disparity maps. We use ground truth maps to eliminate any confusion about the source of error. 
%Without information about the camera, we only know normalized disparity.
We further test our parameter fitting method on ground truth disparity maps, assuming an oracle monocular depth estimation. We took the same 101 object predictions from 3DShapeGen, applied our method on the ground truth normalized disparity maps and merged the initial prediction and projected disparity maps. 
We found a 32\% improvement in f-score at 1 and a 12\% improvement in Chamfer distance. While the magnitude of the improvement is not as large as with the ground truth, there is still a significant improvement. Figure~\ref{tab:synth_results} compares the performance of the baseline 3DShapeGen prediction, the oracle with ground truth depth and the oracle with ground truth disparity. We show qualitative examples in Figure~\ref{fig:qualitative_synthetic}.\\

% \begin{figure}[b]
%     \centering
%     \includegraphics[width=.40\textwidth]{Figures/synth_results.pdf}
%     \caption{Chamfer distances over a set of objects reconstructed from rendered images of ShapeNet chairs using 3DShapeGen~\cite{thai20213d} trained on images only, using our projection method to project ground truth disparities, and using ground truth depth.}
%     \label{fig:synth_results}
% \end{figure}

% \begin{figure}
% \begin{tabular}{|c|c|c|c}
%     \hline
%     &f-score @ 1 & CD\\
%     \hline
%     Baseline (3DShapeGen) &  0.2278 & 0.1064\\
%     \hline
%     GT Depth &0.4938& 0.0640\\
%     \hline
%     GT Disp. Map& 0.2899 & 0.0924 \\
%     \hline
% \end{tabular}
% \caption{Mean f-score and Chamfer distance for the baseline, ground truth depth, and ground truth disparity.}  
% \end{figure}

\subsection{Real Images}
  \begin{figure}[h]
    \centering
    \begin{subfigure}{.45\textwidth}
    \resizebox{\textwidth}{!}{
        \begin{tabular}{|c|c|c|c|}
        \hline
        Method & & f-score & CD \\ 
        \hline
        \multirow{3}{*}{Baseline (Mesh R-CNN)}& Min. & 0.0074 & 0.0680 \\
        & Max. & 0.2857 & 0.7242 \\
        & Mean & 0.1410 & 0.1458 \\
        \hline
        \multirow{3}{*}{Predicted Disparity}& Min. & 0.0068 & 0.0642 \\
        & Max. & 0.3632 & 0.6536 \\
        & Mean & 0.1319 & 0.1476 \\
        \hline
    \end{tabular}}
    \end{subfigure}
    \begin{subfigure}{.5\textwidth}
        \includegraphics[width=\textwidth]{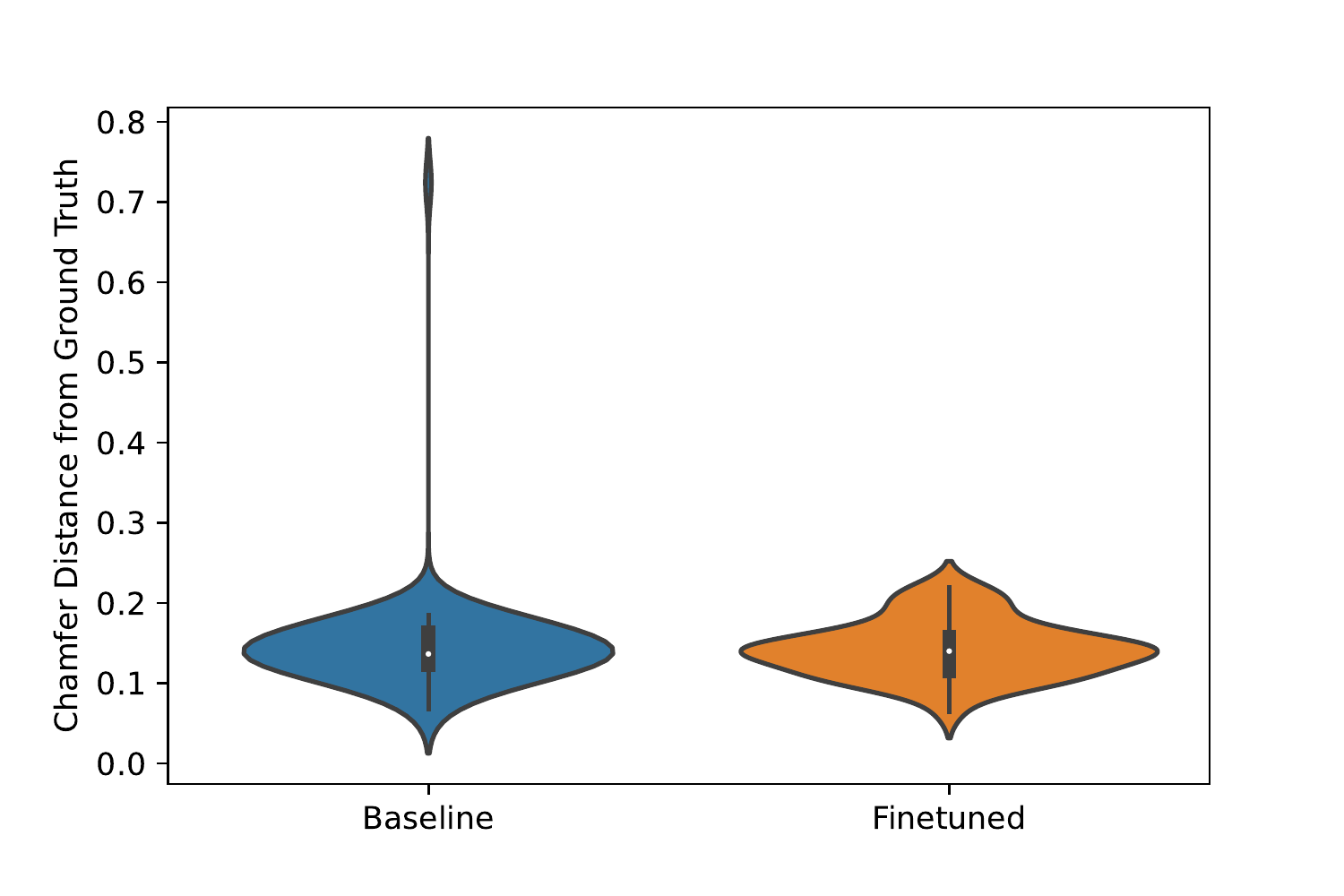}
    \end{subfigure}     

    \caption{We compare the performance of Mesh R-CNN~\cite{gkioxari2019mesh}  with and without our fine-tuning, on real images from the Pix3d dataset~\cite{sun2018pix3d}. We use AdelaiDepth~\cite{Yin_2021_CVPR} to  predict disparities. See section~\ref{subsec:discuss} for a detailed discussion on performance.}
    \label{fig:pix3d_results}
\end{figure}

We proceed to test our method on real images. We use 88 images of chairs from Pix3D \cite{sun2018pix3d} as our dataset and Mesh R-CNN~\cite{gkioxari2019mesh} as our initial method. In this experiment, we use AdelaiDepth \cite{Yin_2021_CVPR} to predict disparity, but use ground truth segmentation masks, again to isolate error. In Figure~\ref{fig:pix3d_results}, we show the distribution of Chamfer distances among the objects in the dataset. After application of our method you can see a slight overall improvement in Chamfer distance.

\begin{figure*}[t]
\resizebox{\textwidth}{!}{
\begin{tabular}{ccccc}
    Input Image & Mesh R-CNN (Baseline) & Baseline Rotated & Fine-Tuning (Ours) & Fine-Tuning Rotated \\
    \includegraphics[width=.15\textwidth]{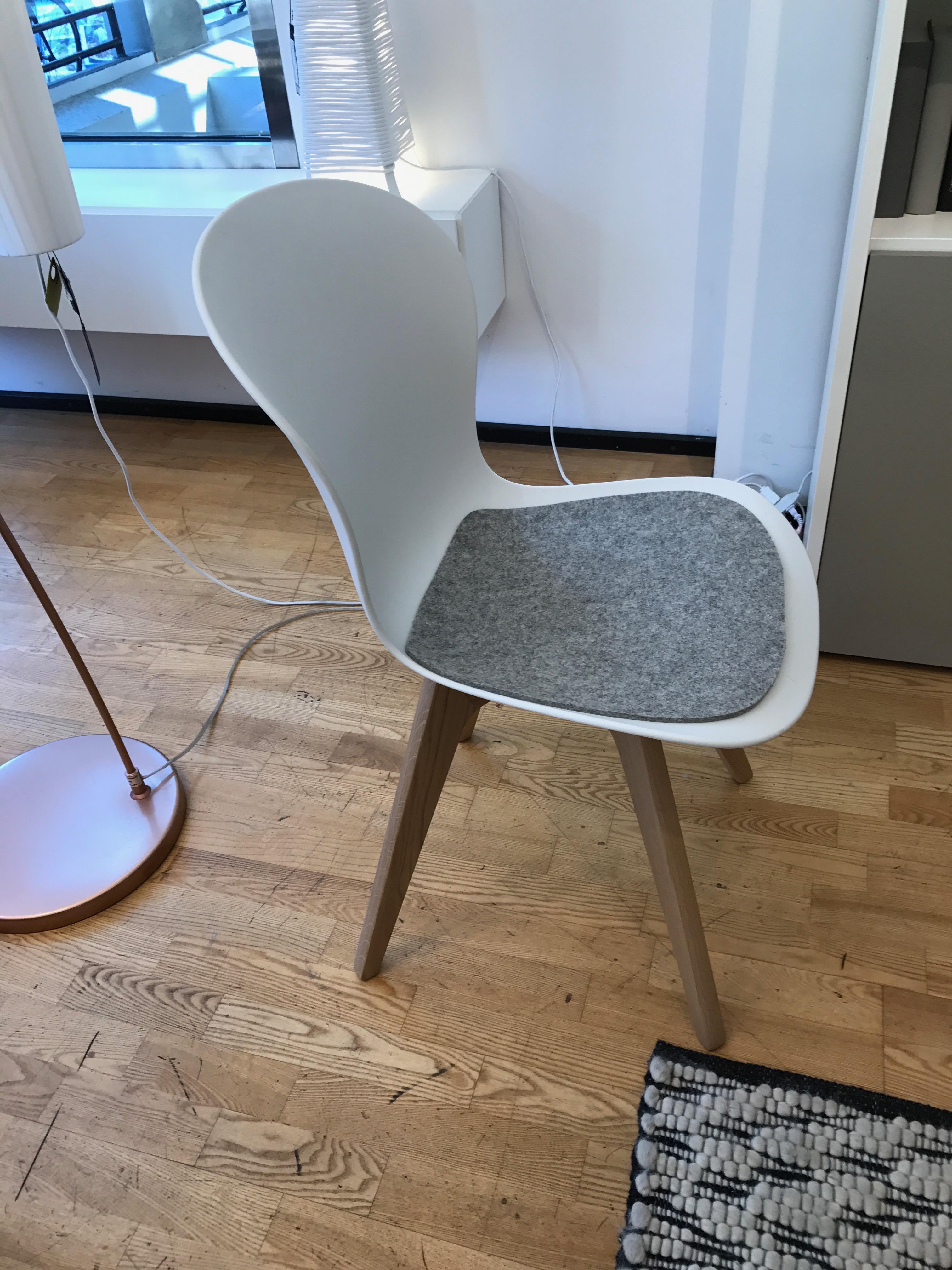} & 
    \includegraphics[width=.15\textwidth]{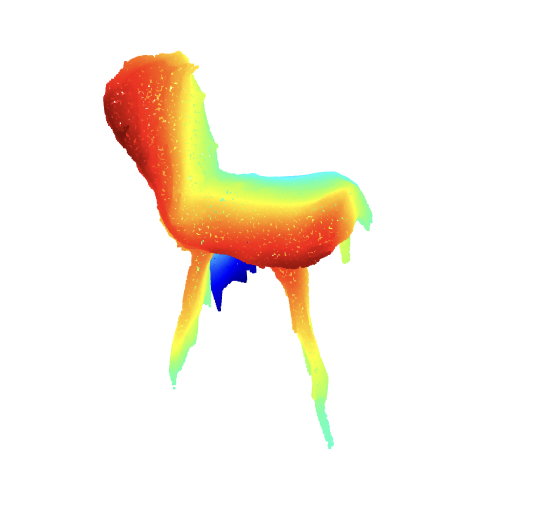} & 
    \includegraphics[width=.15\textwidth]{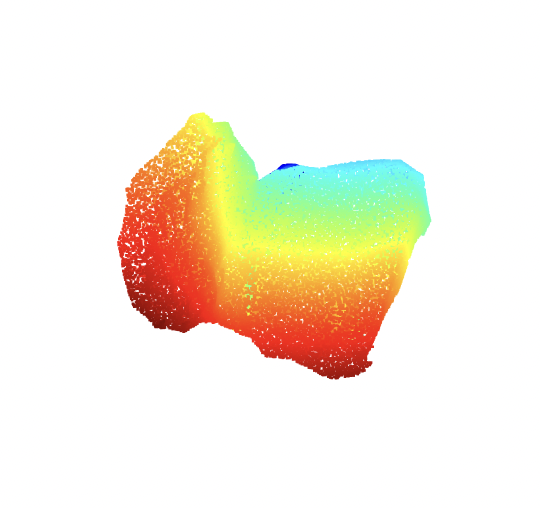} & 
    \includegraphics[width=.15\textwidth]{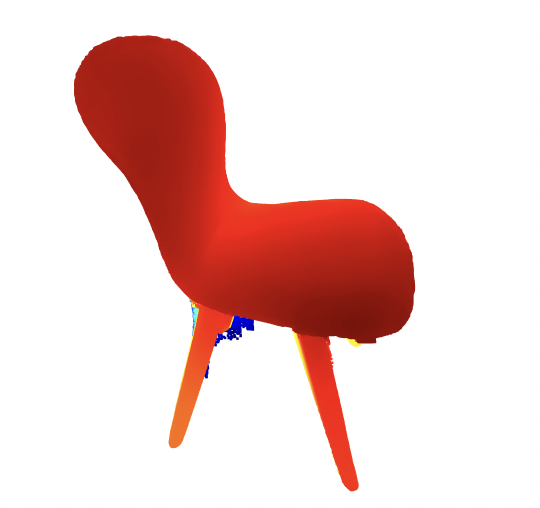} & 
    \includegraphics[width=.15\textwidth]{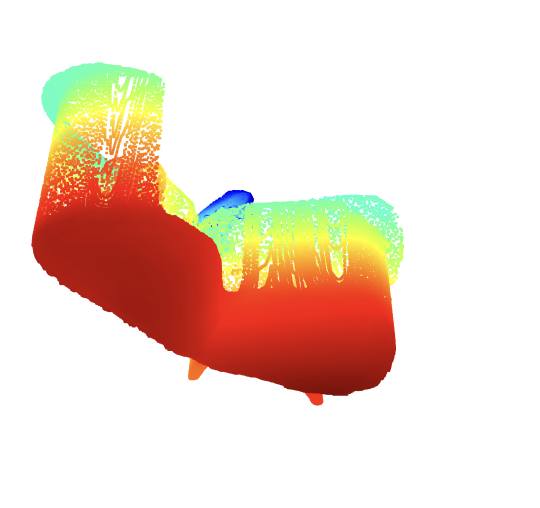}  \\
    
    \includegraphics[width=.15\textwidth]{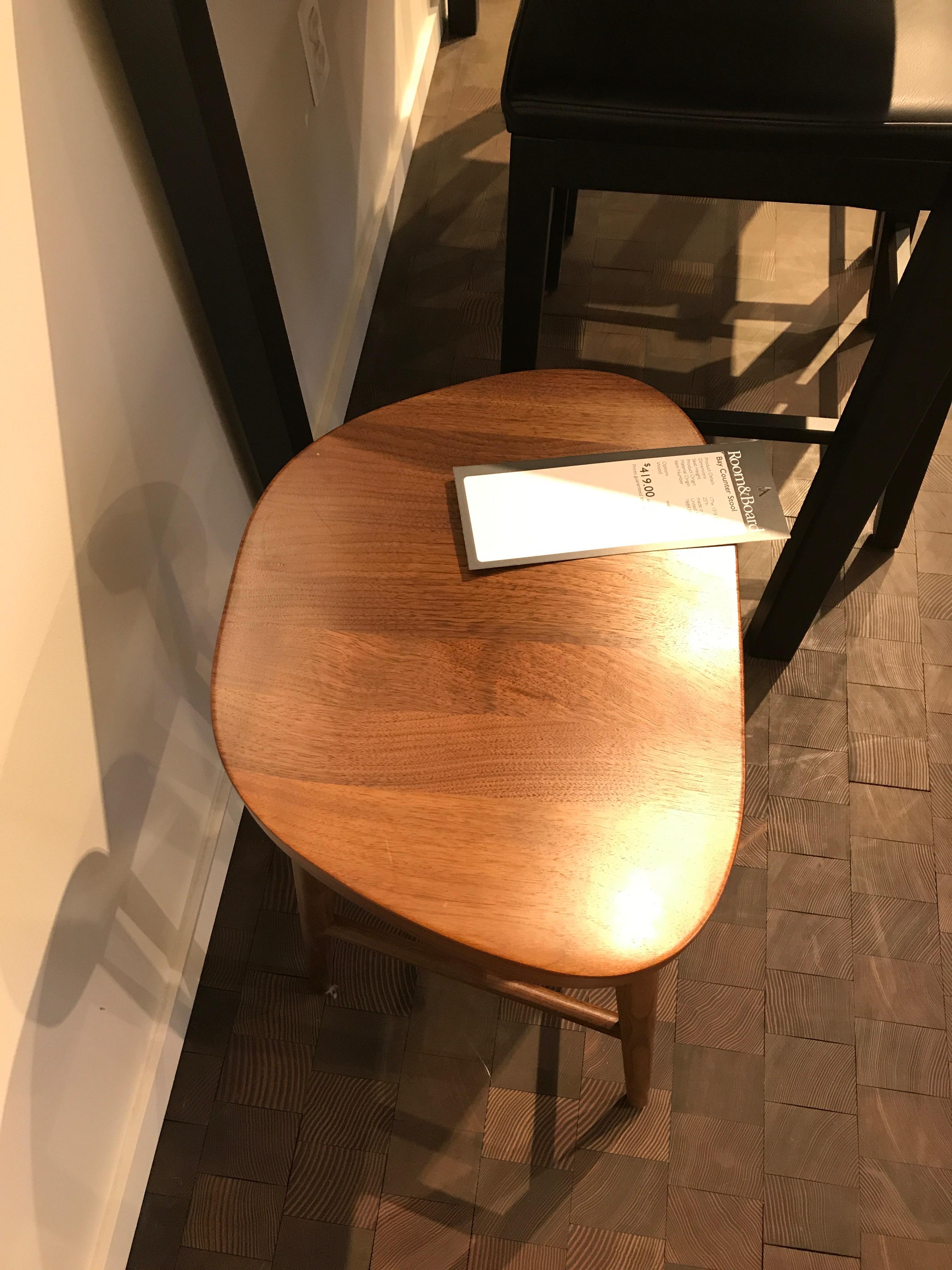} & 
    \includegraphics[width=.15\textwidth]{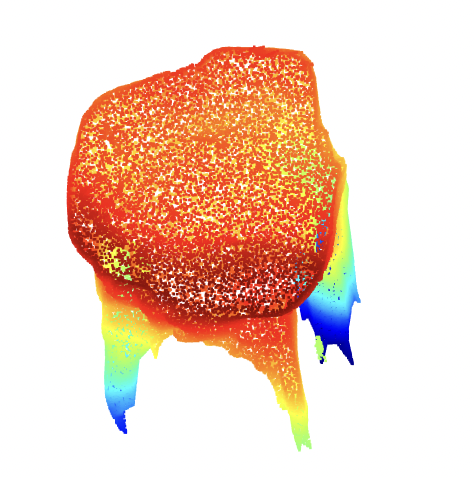} & 
    \includegraphics[width=.15\textwidth]{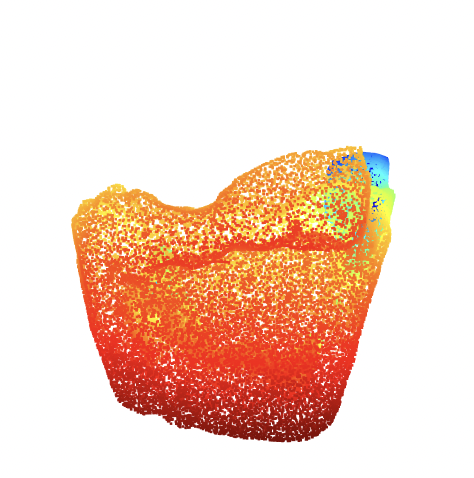} & 
    \includegraphics[width=.15\textwidth]{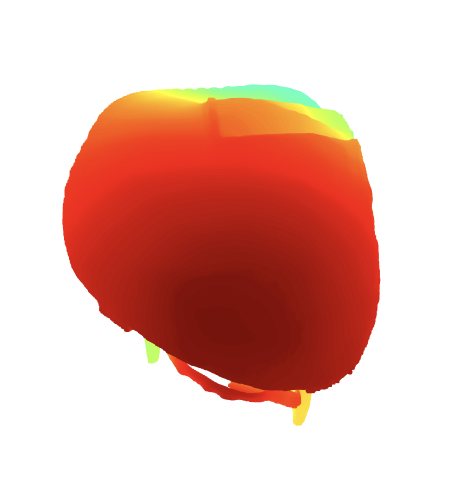} & 
    \includegraphics[width=.15\textwidth]{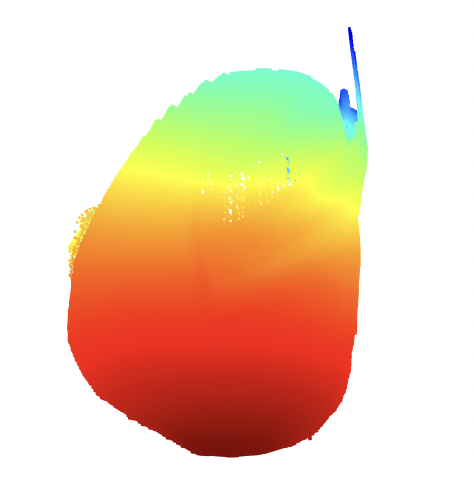} \\
    \includegraphics[width=.15\textwidth]{Figures/0883.jpg} & 
    \includegraphics[width=.15\textwidth]{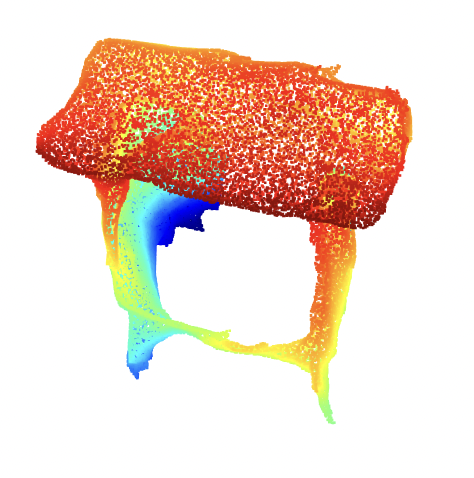} & 
    \includegraphics[width=.15\textwidth]{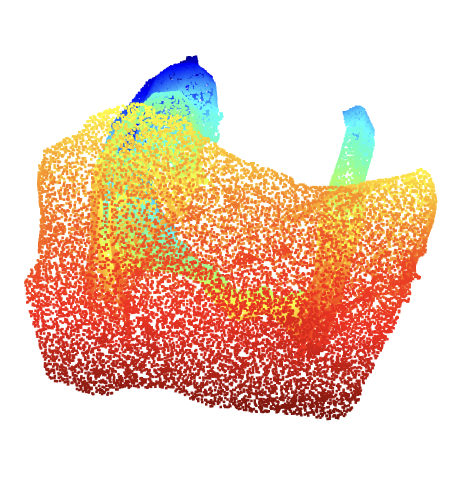} & 
    \includegraphics[width=.15\textwidth]{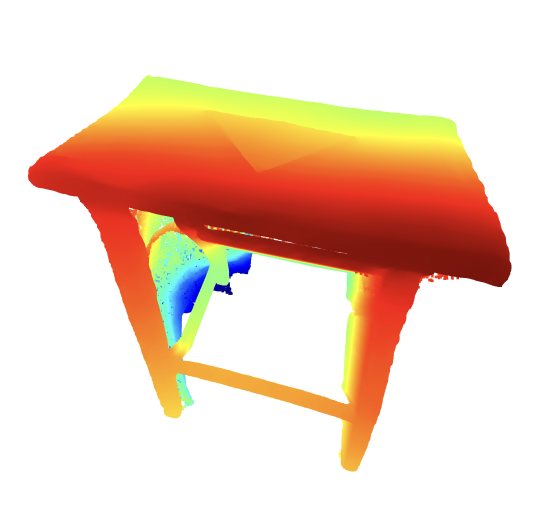} & 
    \includegraphics[width=.15\textwidth]{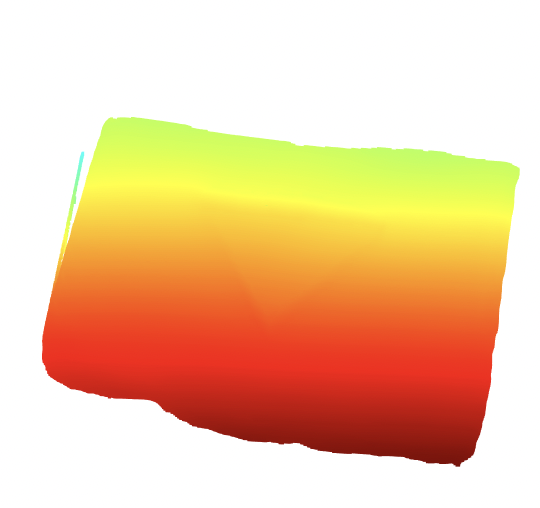} \\
    \includegraphics[width=.15\textwidth]{Figures/0372.jpg} & 
    \includegraphics[width=.15\textwidth]{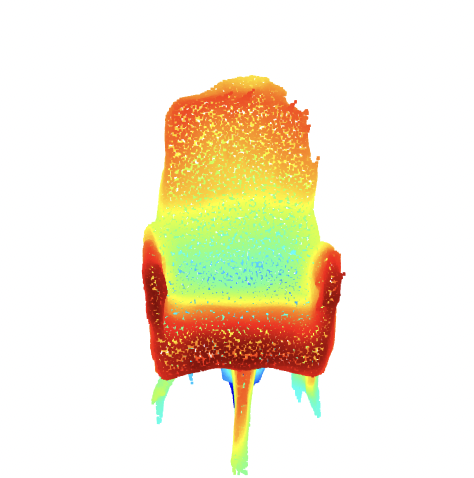} & 
    \includegraphics[width=.15\textwidth]{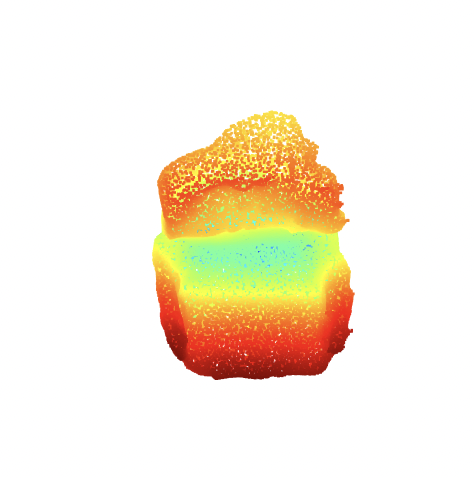} & 
    \includegraphics[width=.15\textwidth]{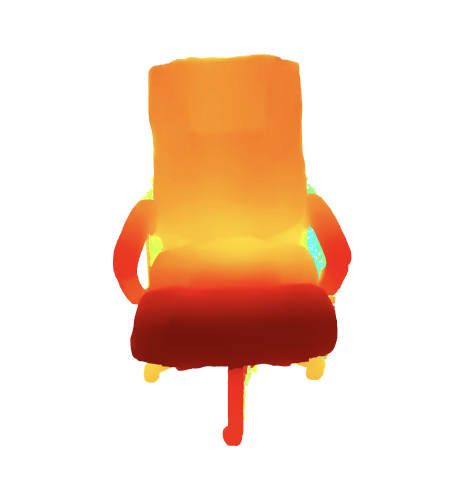} & 
    \includegraphics[width=.15\textwidth]{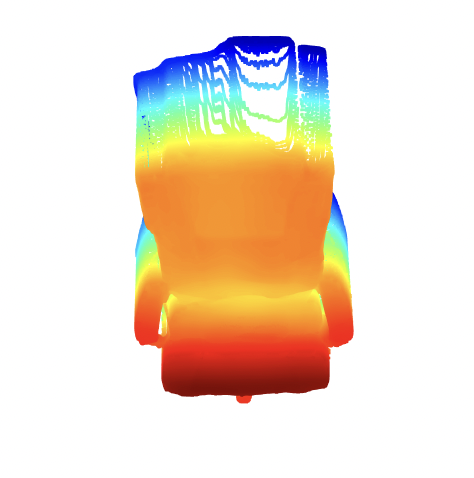} \\
    \includegraphics[width=.15\textwidth]{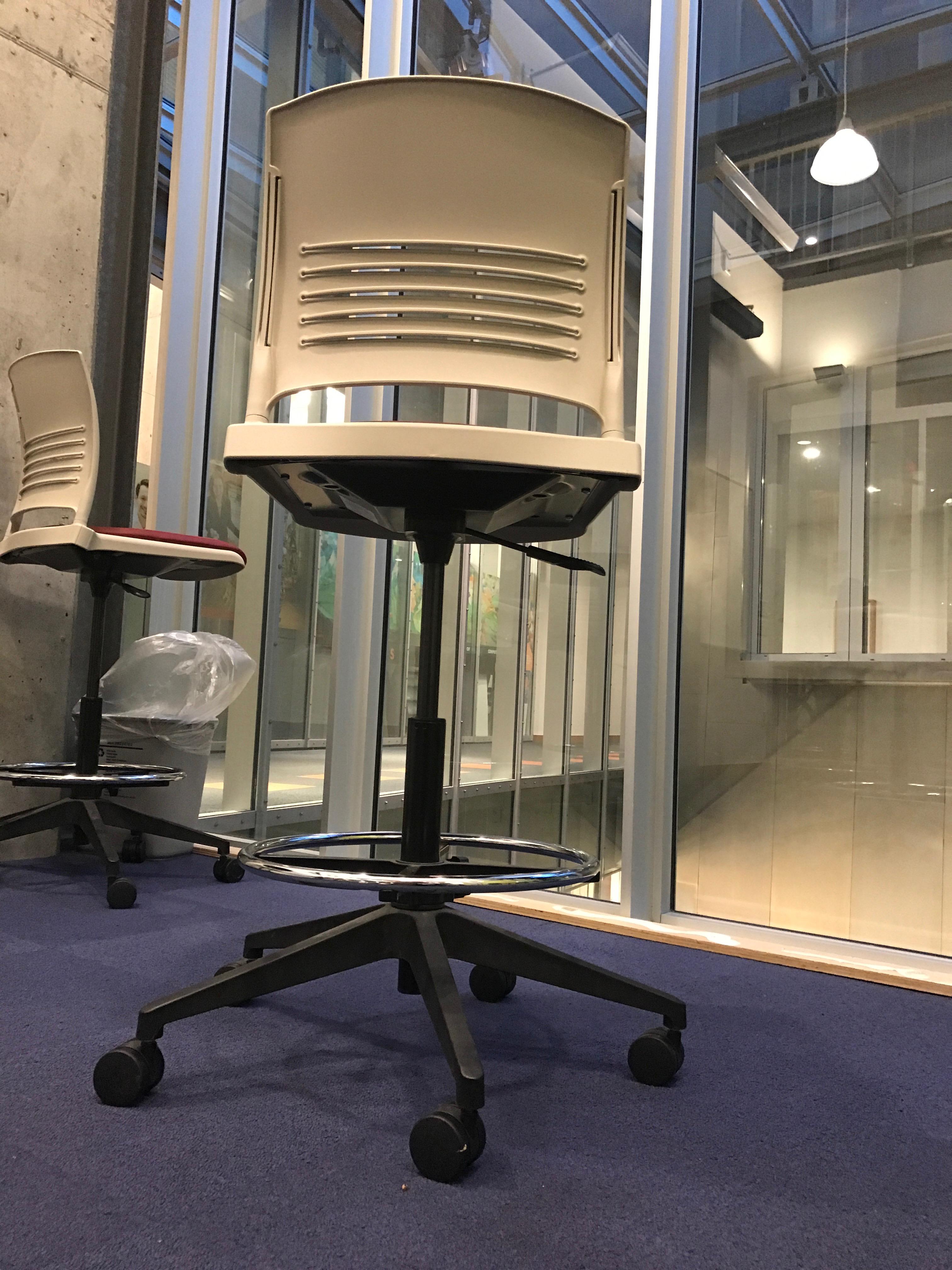} & 
    \includegraphics[width=.15\textwidth]{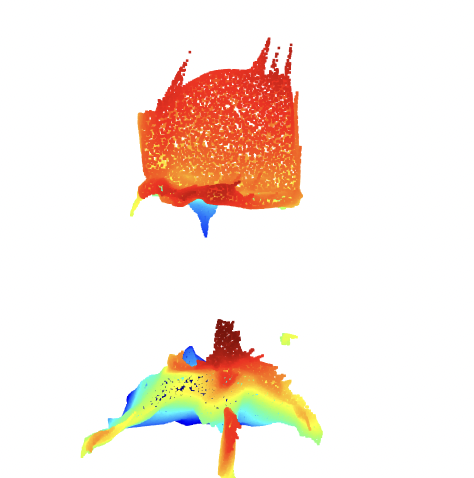} & 
    \includegraphics[width=.15\textwidth]{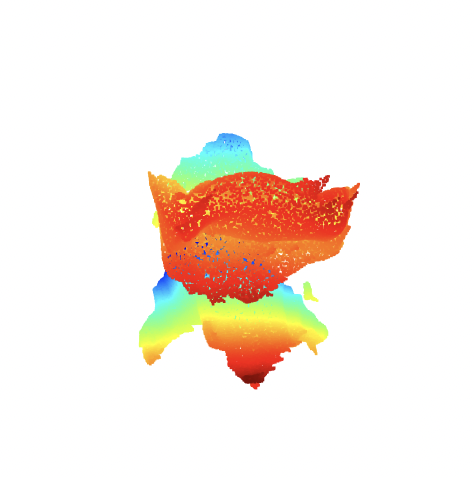} & 
    \includegraphics[width=.15\textwidth]{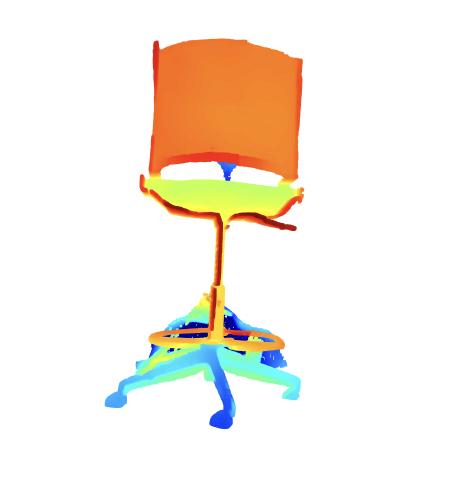} & 
    \includegraphics[width=.15\textwidth]{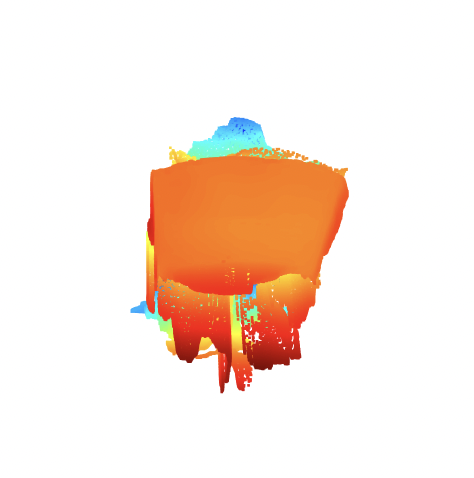} 
    
\end{tabular}}
\caption{Selected qualitative results from the study on real images from Pix3D with estimated normalized disparity.}
\label{fig:qualitative_real}
\end{figure*}

We found that 44\% of objects improved in Chamfer distance from the ground truth with application of our method. In cases that did not see improvement, it was most often from clear errors in the depth estimation. We show qualitative examples in Figure~\ref{fig:qualitative_real}.

\subsection{Discussion}
\label{subsec:discuss}

\begin{figure*}
\resizebox{\textwidth}{!}{
\begin{tabular}{ccccc}
    Input Image & 3DShapeGen (Baseline) & Baseline Rotated & Fine-Tuning (Ours) & Fine-Tuning Rotated \\
    \includegraphics[width=.15\textwidth]{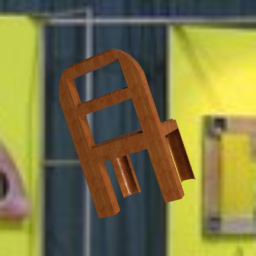} & 
    \includegraphics[width=.15\textwidth]{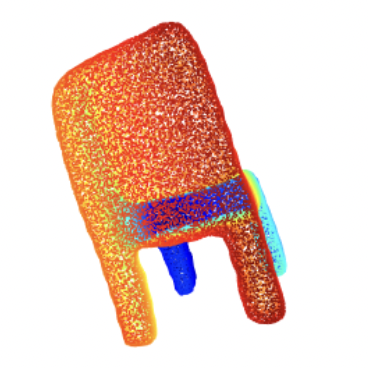} & 
    \includegraphics[width=.15\textwidth]{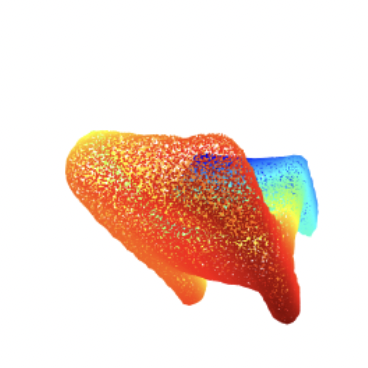} & 
    \includegraphics[width=.15\textwidth]{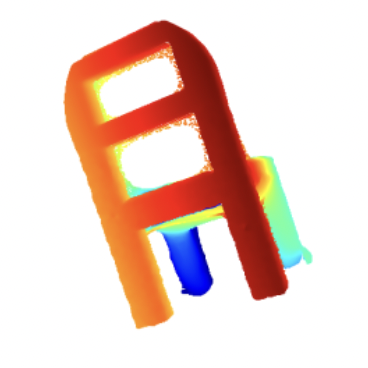} & 
    \includegraphics[width=.15\textwidth]{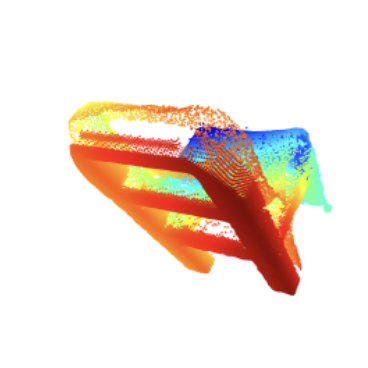}  \\
    
    \includegraphics[width=.15\textwidth]{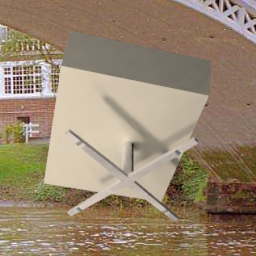} & 
    \includegraphics[width=.15\textwidth]{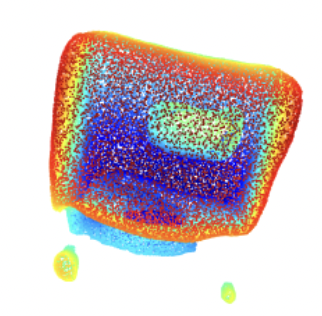} & 
    \includegraphics[width=.15\textwidth]{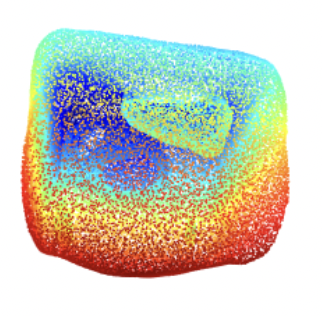} & 
    \includegraphics[width=.15\textwidth]{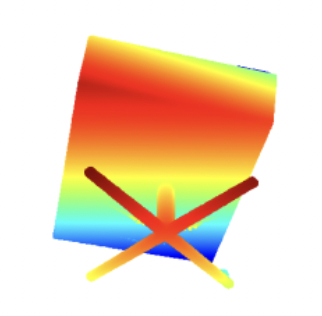} & 
    \includegraphics[width=.15\textwidth]{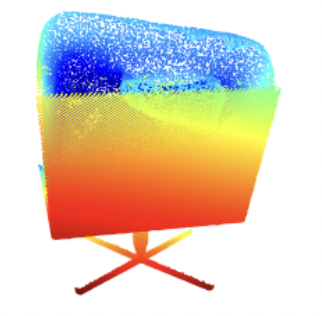} \\
    \includegraphics[width=.15\textwidth]{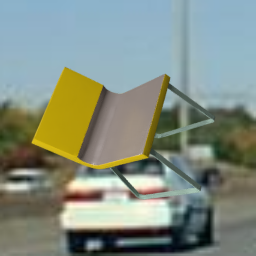} & 
    \includegraphics[width=.15\textwidth]{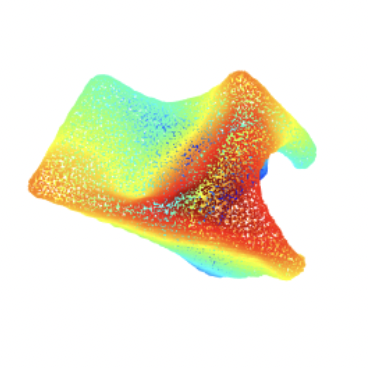} & 
    \includegraphics[width=.15\textwidth]{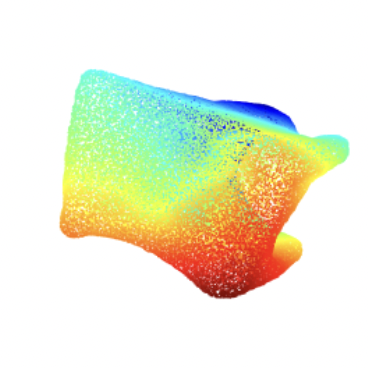} & 
    \includegraphics[width=.15\textwidth]{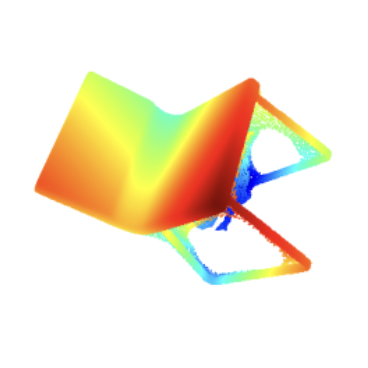} & 
    \includegraphics[width=.15\textwidth]{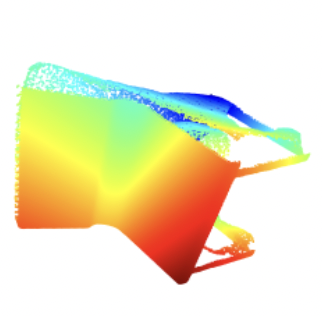} \\
    \includegraphics[width=.15\textwidth]{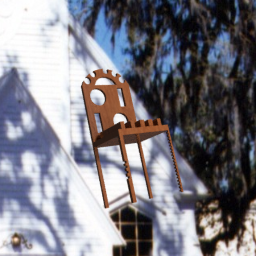} & 
    \includegraphics[width=.15\textwidth]{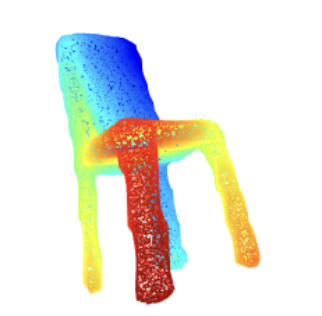} & 
    \includegraphics[width=.15\textwidth]{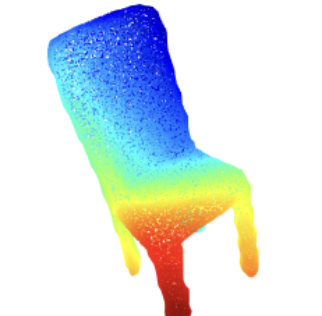} & 
    \includegraphics[width=.15\textwidth]{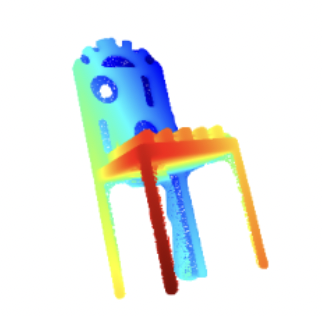} & 
    \includegraphics[width=.15\textwidth]{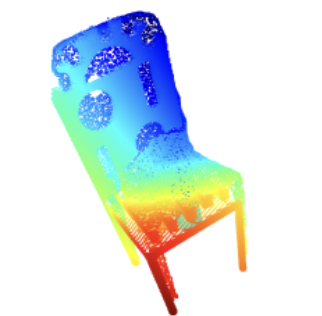} \\
    \includegraphics[width=.15\textwidth]{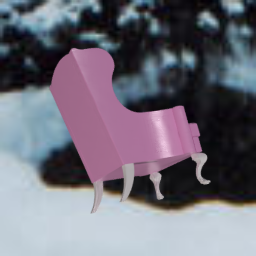} & 
    \includegraphics[width=.15\textwidth]{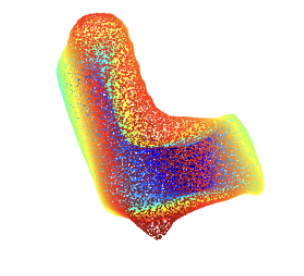} & 
    \includegraphics[width=.15\textwidth]{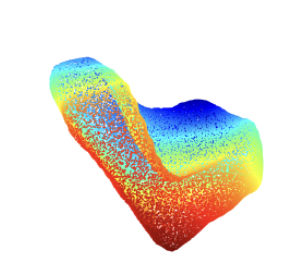} & 
    \includegraphics[width=.15\textwidth]{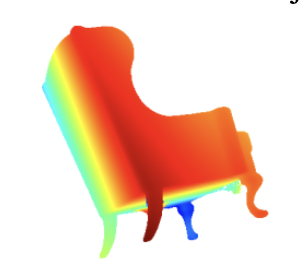} & 
    \includegraphics[width=.15\textwidth]{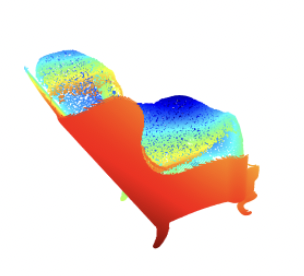} \\
    \includegraphics[width=.15\textwidth]{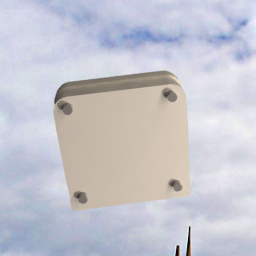} & 
    \includegraphics[width=.15\textwidth]{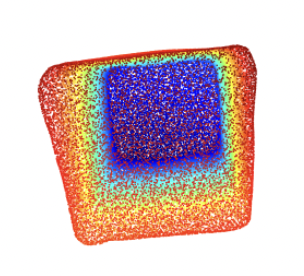} & 
    \includegraphics[width=.15\textwidth]{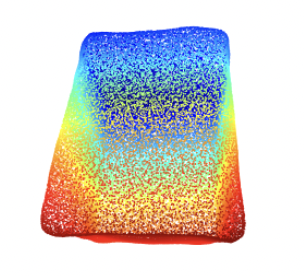} & 
    \includegraphics[width=.15\textwidth]{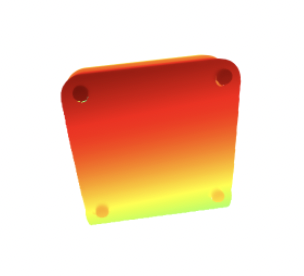} & 
    \includegraphics[width=.15\textwidth]{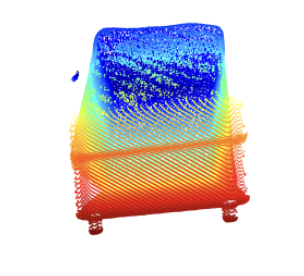} \\
    \includegraphics[width=.15\textwidth]{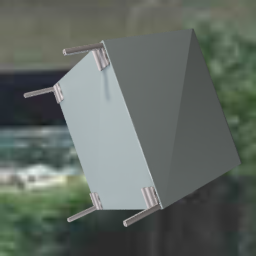} & 
    \includegraphics[width=.15\textwidth]{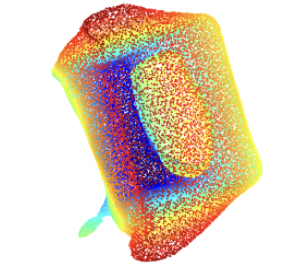} & 
    \includegraphics[width=.15\textwidth]{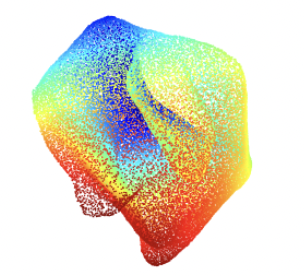} & 
    \includegraphics[width=.15\textwidth]{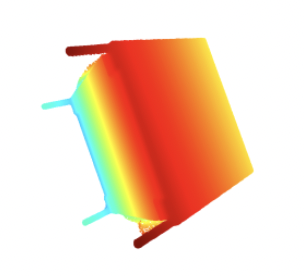} & 
    \includegraphics[width=.15\textwidth]{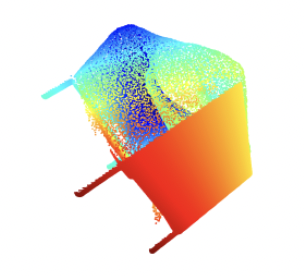} \\
    \includegraphics[width=.15\textwidth]{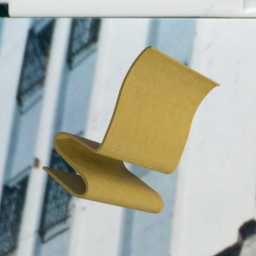} & 
    \includegraphics[width=.15\textwidth]{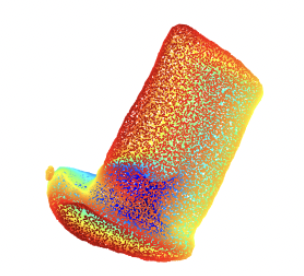} & 
    \includegraphics[width=.15\textwidth]{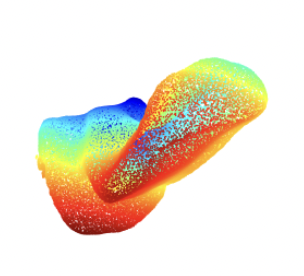} & 
    \includegraphics[width=.15\textwidth]{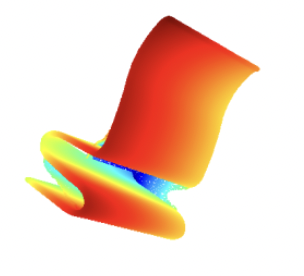} & 
    \includegraphics[width=.15\textwidth]{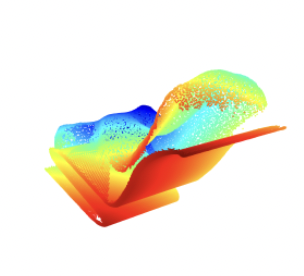} 
\end{tabular}}
\caption{Selected qualitative results from study on synthetic images with ground truth normalized disparity. Our method successfully adds much needed detail to the baseline reconstruction. That detail is preserved not just from the view of the image, but upon rotation as well.}
\label{fig:qualitative_synthetic}
\end{figure*}

The experiments demonstrate the great potential of depth for fine-tuning 3D geometries and the success of our disparity projection approach. In particular, our method is good for recovering fine detail that SVR models often miss. An example is shown in Fig.~\ref{fig:qualitative_synthetic}. Take a look at the chairs in the first and fourth rows. The single-view reconstruction model, reconstructs two fairly similar looking chairs with rounded back and no arms, roughly accurate in shape, but missing most of the detail. Applying our fine-tuning method restores the fine detail that is necessary to distinguish the two chairs. 

Another great use case is fine detail in the legs. Again looking at Fig.~\ref{fig:qualitative_synthetic}, rows two, five, six, and seven are all examples of chairs with very thin, short, or otherwise unusual legs that the single view reconstruction model failed to reconstruct. Our model was able to reconstruct these legs, creating a more complete and distinctive object. 

The objects that fail to see significant improvement in appearance after our fine tuning method tend to fall into two categories: insurmountable errors in the initial prediction and insurmountable errors in depth estimation. 

\begin{figure}[h]
\centering
\begin{subfigure}{0.24\textwidth}
\includegraphics[width=0.7\textwidth]{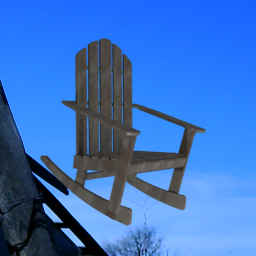}
\caption{Input Image}
\end{subfigure}
\begin{subfigure}{0.24\textwidth}
\includegraphics[width=\textwidth]{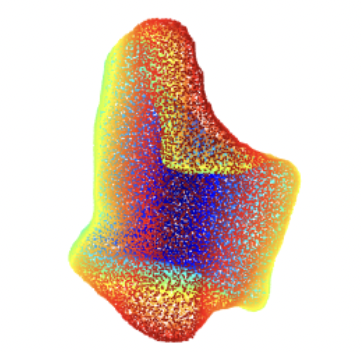}
\caption{Baseline}
\end{subfigure}
\begin{subfigure}{0.48\textwidth}
\includegraphics[width=0.5\textwidth]{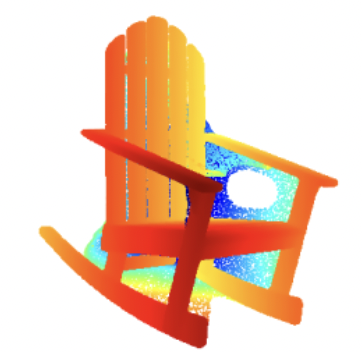}\includegraphics[width=0.5\textwidth]{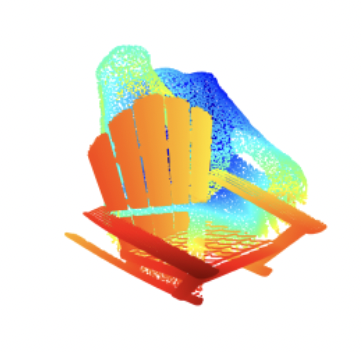}
\caption{Fine-Tuning (2 views)}
\end{subfigure}

\caption{An example of a flattened projection due to errors in the initial prediction.}
\label{fig:flattened}
\end{figure}

 Fig.~\ref{fig:flattened} shows a typical example of an insurmountable error in the initial prediction, where the network creates a blobby initial prediction. Because of this blobby shape, the visible points of the initial prediction are mostly flat and the estimated disparity is thus projected to a flat point cloud even though the actual visible pixels are not flat.

\begin{figure}
    \centering
    \begin{subfigure}{0.35\textwidth}
    \includegraphics[width=0.7\textwidth]{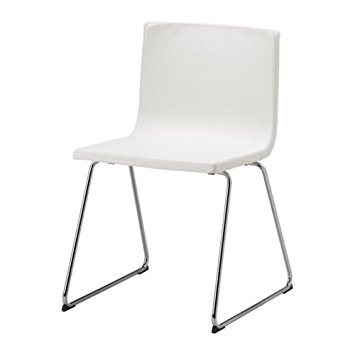}
    \caption{Input Image}
    \end{subfigure}
    \begin{subfigure}{0.35\textwidth}
    \includegraphics[width=0.7\textwidth]{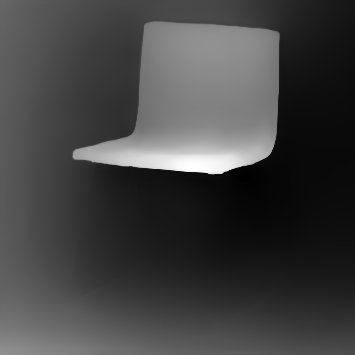}
    \caption{Predicted Disparity}
    \end{subfigure}
    \caption{An insurmountable mistake in disparity prediction: the legs a treated as background.}
    \label{fig:disp_errors}
\end{figure}

The second major source of errors lies in the disparity estimation. Our method is fairly robust to minor errors which will occur in any depth estimation method, but there are large fundamental errors that are insurmountable with our approach. An example is the chair in Fig.~\ref{fig:disp_errors} where the monocular depth estimation failed to detect the legs of the chair, so they blend in with the background. Even with perfect segmentation and camera parameter estimation, the legs of the chair would be placed in the background away from the chair. These types of errors account for most of the objects in our dataset that failed to improve, as evidenced by the difference in performance between the ground truth normalized disparity maps and the predicted ones. 
As monocular depth estimation continues to improve, these errors will become less common.

\section{Conclusion}

In summary, we have demonstrated how to improve the predictions of single-view reconstruction models using monocular-depth estimation. We presented a method to approximate the relevant camera parameters to convert normalized disparity, from an off-the-shelf monocular depth model, to a 3D point cloud. We tested our method via multiple experiments, quantified and analysed the improvement that disparity maps can provide on real images.

\section*{Acknowledgement}
This research used resources of the National Energy Research Scientific Computing Center (NERSC), a U.S. Department of Energy Office of Science User Facility located at Lawrence Berkeley National Laboratory, operated under Contract No. DE-AC02-05CH11231.

\bibliographystyle{splncs}
\bibliography{bib}
\end{document}